\documentclass[lettersize,journal]{IEEEtran}
\usepackage{amsmath,amsfonts}
\usepackage[T1]{fontenc}
\usepackage{rotating}
\usepackage{array}
\usepackage{textcomp}
\usepackage[ruled,vlined,onelanguage]{algorithm2e} 
\usepackage{algpseudocode}
\usepackage[table,xcdraw]{xcolor}
\usepackage{stfloats}
\usepackage{url}
\usepackage{verbatim}
\usepackage{graphicx}
\usepackage{longtable}
\usepackage{academicons}
\usepackage{multirow}
\newcommand{\orcid}[1]{\href{https://orcid.org/#1}{\textcolor[HTML]{A6CE39}{\aiOrcid}}}
\usepackage{tikz}
\usetikzlibrary{svg.path}
\usepackage[colorlinks=true, pdfstartview=FitV, linkcolor=blue, citecolor=blue, urlcolor=blue]{hyperref}
\usepackage{lscape}
\usepackage{cite}
\usepackage{amsmath}
\usepackage{amssymb}
\usepackage{xcolor}
\usepackage{caption}
\usepackage{xr}
\usepackage{orcidlink}
\usepackage{booktabs,caption}
\usepackage[flushleft]{threeparttable}

\usepackage[colorlinks=true, pdfstartview=FitV, linkcolor=blue, citecolor=blue, urlcolor=blue]{hyperref}

\begin{document}

\title{Enhancing Trajectory Prediction
through Self-Supervised Waypoint Noise Prediction}

\author{Pranav Singh Chib$^{\orcidlink{0000-0003-4930-3937}}$, \and Pravendra Singh$^{\orcidlink{0000-0003-1001-2219}}$


\thanks{(Corresponding author: Pravendra Singh.)}
\thanks{Pranav Singh Chib and Pravendra Singh are with the Department of Computer Science and Engineering, Indian Institute of Technology Roorkee, Uttarakhand 247667, India,
(e-mail: pranavs\_chib@cs.iitr.ac.in; pravendra.singh@cs.iitr.ac.in).}}



\maketitle

\begin{abstract}

Trajectory prediction is an important task that involves modeling the indeterminate nature of traffic actors to forecast future trajectories given the observed trajectory sequences. However, current methods confine themselves to presumed data manifolds, assuming that trajectories strictly adhere to these manifolds, resulting in overly simplified predictions. To this end, we propose a novel approach called SSWNP (Self-Supervised Waypoint Noise Prediction). In our approach, we first create clean and noise-augmented views of past observed trajectories across the spatial domain of waypoints. We then compel the trajectory prediction model to maintain spatial consistency between predictions from these two views, in addition to the trajectory prediction task. Introducing the noise-augmented view mitigates the model's reliance on a narrow interpretation of the data manifold, enabling it to learn more plausible and diverse representations. We also predict the noise present in the two views of past observed trajectories as an auxiliary self-supervised task, enhancing the model's understanding of the underlying representation and future predictions. Empirical evidence demonstrates that the incorporation of SSWNP into the model learning process significantly improves performance, even in noisy environments, when compared to baseline methods. Our approach can complement existing trajectory prediction methods. To showcase the effectiveness of our approach, we conducted extensive experiments on three datasets: NBA Sports VU, ETH-UCY, and TrajNet++, with experimental results highlighting the substantial improvement achieved in trajectory prediction tasks.
\end{abstract}


\section{Introduction}
\IEEEPARstart{T}{rajectory} prediction involves estimating an agent's future motions by analyzing their historical past trajectories. This process holds significant importance in various applications, including autonomous driving, robotics, surveillance systems, drones, and other autonomous systems. The future trajectories of agents (such as pedestrians and vehicles) often exhibit uncertainty due to the ability of agents to adapt their movement in response to changing environments and physical constraints. Given observed past trajectories, multiple potential future paths exist for agents. Consequently, an effective motion forecasting method should be capable of generating a distribution of potential future trajectories or, at the very least, several probable ones.

Several research studies have focused on utilizing deep generative models \cite{mao2023leapfrog, Xu_2022_CVPR, gu2022stochastic}. For example, some approaches employ generative adversarial networks (GANs) \cite{sophie19, hu2020collaborative, gupta2018social} to diversify the distribution across all potential future trajectories. In contrast, alternative approaches \cite{Xu_2022_CVPR, lee2022muse, xu2022dynamic, mangalam2020not} utilize conditional variational autoencoders (CVAE) to capture the multi-modal distribution of future trajectories. Transformer models \cite{giuliari2021transformer, tsao2022social, girgis2022latent, vip3d} better learn spatial and temporal dependencies. Graph-based models \cite{Xu_2022_CVPR, xu2023uncovering, bae2023set, lv2023ssagcn, sekhon2021scan} have taken a step forward in modeling the complex social interaction and the uncertain nature of trajectories. Despite receiving significant attention and featuring various proposed architectures, trajectory prediction models may encounter the challenge of overly simplified predictions. This issue persists due to the model's reliance on a narrow interpretation of the data manifold. When there are no distinct movement patterns among most trajectories, the prediction model tends to generate simple or uniform trajectories, failing to capture variations in the motion patterns of entities such as pedestrians or vehicles. This could lead to less accurate predictions, especially in scenarios where diverse and complex trajectory patterns must be considered.

\begin{figure*}
    \centering
    \includegraphics[scale=.1]{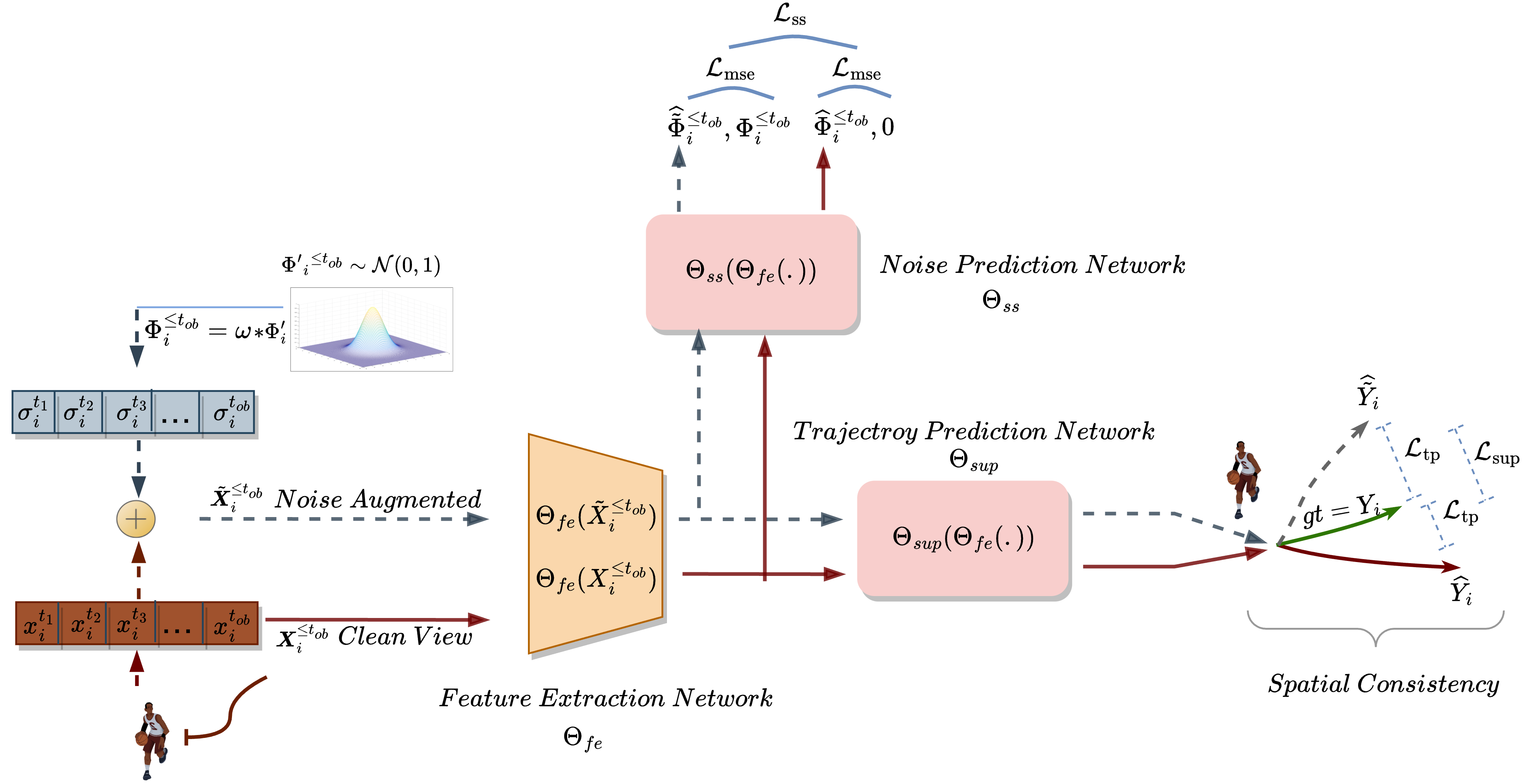}
    \caption{Illustration of our SSWNP, in which we first create two different views of past observed trajectories: one as the clean view ${\boldsymbol{X}}_{i}^{\leq t_{ob}}$ and the other as a noise-augmented view $\tilde{\boldsymbol{X}}_{i}^{\leq t_{ob}}$ across the spatial domain of waypoints. We then compel the trajectory prediction model $\Theta_{sup}$ to maintain spatial consistency between predictions from these two views and predict the future trajectories ${\widehat Y}_{i}$ and ${\widehat{\tilde{Y}}}_{i}$ (Section~\ref{sec_spatial_cons}). In the noise prediction module, we predict the noise present in the two views of past observed trajectories as an auxiliary self-supervised task (Section~\ref{sec_noisepred}).}
    \label{first_fig}
\end{figure*}

To address the aforementioned issues, we propose a novel approach called SSWNP (Self-Supervised Waypoint Noise Prediction), which consists of two modules (spatial consistency module and noise prediction module) as shown in Figure~\ref{first_fig}. In the spatial consistency module, we create two different views of past observed trajectories: one as the clean view and the other as a noise-augmented view across the spatial domain of waypoints. As the name suggests, the clean view represents the original past observed trajectory, while the noise-augmented view represents past observed trajectories that are spatially relocated with some additive noise. Our approach leverages the fact that a noise-augmented view of past observed trajectories does not adhere to the narrow interpretation of the data manifold. The model uses this additional information to overcome the challenge of overly simplified predictions and learn more plausible and diverse representations (see Figure \ref{nba_vis}). After creating two different views of past observed trajectories, we compel the trajectory prediction model to maintain spatial consistency between predictions from these two views and learn the spatiotemporal characteristics, in addition to the trajectory prediction task. 

In the noise prediction module, we predict the noise present in the two views of past observed trajectories as an auxiliary self-supervised task, enhancing the model's understanding of the underlying representation. Self-supervision \cite{bhattacharyya2023ssl} has garnered significant attention, aiming to leverage available data without necessitating annotations. Its objective is to assist models in acquiring more generalized representations through pretext tasks. Given the recent achievements in self-supervised learning \cite{wei2019iterative}, we focus on utilizing self-supervised learning to optimize trajectory prediction. In our approach, we propose a novel pretext task of noise prediction in past observed trajectories. This novel self-supervised auxiliary task helps the trajectory prediction model better model the potential spatial diversity and improves the understanding of the underlying representation in trajectory prediction, thereby enhancing future predictions (see Figure \ref{nba_vis}). We also conduct ablation experiments in Section~\ref{ab_module} to empirically demonstrate that both modules (i.e., spatial consistency module and noise prediction module) are crucial for our approach. If we only employ the spatial consistency module alongside the trajectory prediction task, suboptimal performance is observed. Therefore, in our approach, we incorporate both modules along with the trajectory prediction task.

Our approach can be easily integrated with existing trajectory prediction methods. We have integrated our approach into four existing methods: the generative-based GroupNet \cite{Xu_2022_CVPR}, the goal-oriented Graph-TERN \cite{bae2023set}, Graph-based SSAGCN \cite{lv2023ssagcn} and the transformer-based AutoBot \cite{girgis2022latent}. Our extensive experiments demonstrate the ability of our approach to accurately forecast future trajectories, leading to substantial performance improvements across the NBA SportVU \cite{zhan2018generating}, Trajnet++ \cite{Kothari2020HumanTF}, and ETH-UCY \cite{pellegrini2009you} datasets. Additionally, we conduct ablation experiments (see Section~\ref{ab_noisyenv}) to demonstrate that incorporating SSWNP into the model learning process significantly improves performance in a noisy environment compared to the baseline method. 

The major contributions of our work are summarized below.

\begin{itemize}
    \item We propose a novel approach called SSWNP (Self-Supervised Waypoint Noise Prediction), which comprises a spatial consistency module and a noise prediction module, to overcome the challenge of overly simplified predictions.
    \item We propose a novel pretext task of noise prediction in past observed trajectories. This self-supervised auxiliary task helps the trajectory prediction model in improving its understanding of the underlying representation in trajectory prediction, thereby enhancing future predictions.
    \item Our approach can complement existing trajectory prediction methods. We empirically demonstrate that incorporating SSWNP into the model learning process significantly improves performance, even in noisy environments, compared to baseline methods.
\end{itemize}

\section{Related Work}
\label{sec:related}

\subsection{Trajectory prediction}  
The trajectory forecasting model seeks to predict future trajectories considering the observed trajectories. There is inherent uncertainty when forecasting an agent's future trajectory, which usually leads to a wide range of possible trajectories. To model this uncertainty, the stochastic prediction model has been used in several works \cite{shi2021sgcn,gupta2018social,lee2022muse,xu2022dynamic,mangalam2020not}. These models include a range of methods, such as conditional variational autoencoders (CVAEs) \cite{Xu_2022_CVPR,lee2022muse,xu2022dynamic,mangalam2020not}, generative adversarial networks (GANs) \cite{sophie19, hu2020collaborative,gupta2018social}, and diffusion models \cite{mao2023leapfrog,gu2022stochastic}.  Despite significant advancements, these stochastic prediction methods exhibit inherent limitations, such as unstable training or the generation of unnatural trajectories. Some work including RMB \cite{shi2023representing} addresses the issue of superfluous interactions by proposing the Interpretable Multimodality Predictor (IMP), which models the distribution of mean locations as a Gaussian Mixture Model (GMM) and encourages multimodality by sampling multiple mean locations of predictions. Stimulus verification \cite{sun2023stimulus}  proposes an explicit sampling process to improve the final prediction results by sampling highly stimulus-coherent trajectories. Transformer-based \cite{girgis2022latent,yuan2021agentformer,zhou2023query,yu2020spatio} models are widely employed to capture temporal and social dimensions through the attention mechanism concurrently. Moreover, they excel in capturing long-range dependencies. VIKT \cite{zhong2023visual} incorporates visual localization and orientation to enhance trajectory prediction by learning from real-world visual settings. They also leverage Visual Intention Knowledge (VIK) with the spatiotemporal Transformer (VIKT) to represent human intent. VNAGT \cite{chen2023vnagt} introduces a variational non-autoregressive graph transformer to capture social and temporal interactions. LSSTA \cite{yang2023long} proposes a spatial transformer that effectively models the dynamic nature of pedestrian interactions while also accounting for time-varying spatial dependencies. There exists the accumulation error while predicting future trajectories. To mitigate the accumulation of prediction errors, SIM \cite{li2023synchronous} introduces a synchronous bi-directional structure.  Similarly, STS LSTM \cite{zhang2023spatial} models spatiotemporal interactions using LSTM-based architectures. Graph-based \cite{lv2023ssagcn,sekhon2021scan,pmlr-v80-kipf18a} methods are specifically utilized to explicitly model social interactions among agents in the scene through relational reasoning. They adeptly capture interactions and their associated strengths in both groupwise and pairwise interactions to predict plausible future trajectories. DynGroupNet \cite{xu2023dynamic} and TDGCN \cite{wang2023trajectory} focus on capturing temporal groupwise interactions, considering interaction strength and interaction category. Additionally, diverse perspectives have focused on trajectory prediction, such as endpoint-conditioned trajectory prediction \cite{bae2023set}, long-tail trajectory prediction \cite{wang2023fend}, and others.  SRGAT \cite{chen2023goal} incorporates multiple goals predicted for each agent, followed by social interaction modeling. MERA \cite{sun2023modality} utilizes different types of modalities in motion predictions, processing different feature clusters to represent modalities such as scene semantics and agent motion state. The Multi-Style Network (MSN) \cite{wong2023msn} incorporates style as a factor in predictions, providing trajectories with multi-style predictions.  MetaTraj \cite{shi2023metatraj} provides sub-tasks and a meta-task for trajectory prediction that can accommodate predictions for unseen scenes and objects. Additionally, DISTL \cite{cao2023discovering} defines a set of spatial-temporal logic rules to describe human actions.

\subsection{Self-supervised Learning}

Self-supervised learning is a paradigm that has gained popularity across various domains of deep learning, including computer vision. Through different pretext tasks, additional supervision is generated from unlabeled data, which is then used to train a model in a self-supervised manner. Several self-supervised approaches \cite{gidaris2018unsupervised, wei2019iterative, caron2018deep, pathak2017learning} have been developed to acquire better representation learning. For example, in the context of acquiring image features \cite{gidaris2018unsupervised}, self-supervised tasks train deep networks to recognize the 2D rotation angles of images. In another approach \cite{wei2019iterative}, a pretext task is proposed to learn spatial relationships by dividing an image into a grid of patches, rearranging their spatial positions, and training the network to restore their accurate spatial arrangement. Additional self-supervised learning techniques include image clustering \cite{caron2018deep}, segmentation prediction \cite{pathak2017learning}, and others. These tasks assist the model in learning the underlying representation.

Recently, in trajectory prediction \cite{bhattacharyya2023ssl, wang2023fend, halawa2022action}, a few works have explored self-supervised learning. Some employ contrastive learning \cite{wang2023fend, halawa2022action} to enhance the representation ability of the network, while others, like SSL lanes \cite{bhattacharyya2023ssl}, utilize map/agent-level data to formulate various pretext tasks. Unlike the above-mentioned methods, we propose a novel pretext task that predicts the noise present in the clean and noise-augmented views of past observed trajectories as an auxiliary self-supervised task to enhance the trajectory prediction task.

\subsection{Learning with Regularization}
Several techniques have been explored in recent studies to regularize trajectory prediction. Some methodologies, like the one proposed by Ye et al. \cite{ye2022bootstrap}, utilize a variety of transformations applied to the same input data to generate perturbation-invariant representations. This approach emphasizes temporal consistency, ensuring that inputs undergoing slight time interval shifts produce similar output trajectories. Wu et al. \cite{wu2023masked} employed masked trajectory predictions and reconstruction to extract additional signals from trajectories. TENET \cite{wang2022tenet} propagates learning embeddings through a temporal flow network to reconstruct the input, serving as a means to enhance the acquired embeddings. Researchers \cite{girgis2022latent, sekhon2021scan, zhu2020robust} also implement a mechanism wherein the predicted future trajectory is reversed temporally and fed back into the prediction model. This approach aims to predict the historical trajectory, and the loss is calculated with the inclusion of an extra cycle loss term. This prolongs their training process since they need to undergo additional training. In contrast to the above-mentioned methods, our approach maintains spatial consistency between predictions from clean and noise-augmented views of past observed trajectories across the spatial domain of waypoints, in addition to the trajectory prediction task.

\section{Methodology}
\label{sec:meth}

\subsection{Problem Formulation}

The goal of trajectory predictions is to forecast the future trajectories of agents in a dynamic environment based on their past trajectories. A trajectory is represented by a temporal series of spatial points, termed as waypoints. The past observed trajectory, spanning from $t_{1}$ to $t_{ob}$, can be denoted as $X_{i}^{\leq t_{ob}}=\{\boldsymbol{x}_{i}^{t_1}, \boldsymbol{x}_{i}^{t_2},..., \boldsymbol{x}_{i}^{t_{ob}}\}$, where $\boldsymbol{x}_{i}^{t_{ob}} \in \mathbb{R}^{2}$ 
corresponds to the 2D coordinates of agent $i$ at time step $t_{ob}$. Similarly, the predicted future trajectory for agent $i$ over the duration $t_{ob+1}$ to $t_{fu}$ can be described as ${\widehat Y}_{i}^{t_{ob+1} \leq t \leq t_{fu}}$. Corresponding ground truth for the future trajectory of agent $i$ can be described as $ Y_{i}^{t_{ob+1} \leq t \leq t_{fu}}$ over the duration $t_{ob+1}$ to $t_{fu}$.

\subsection{Self-Supervised Waypoint Noise Prediction}

\begin{figure}
    \centering
    \includegraphics[scale=.42]{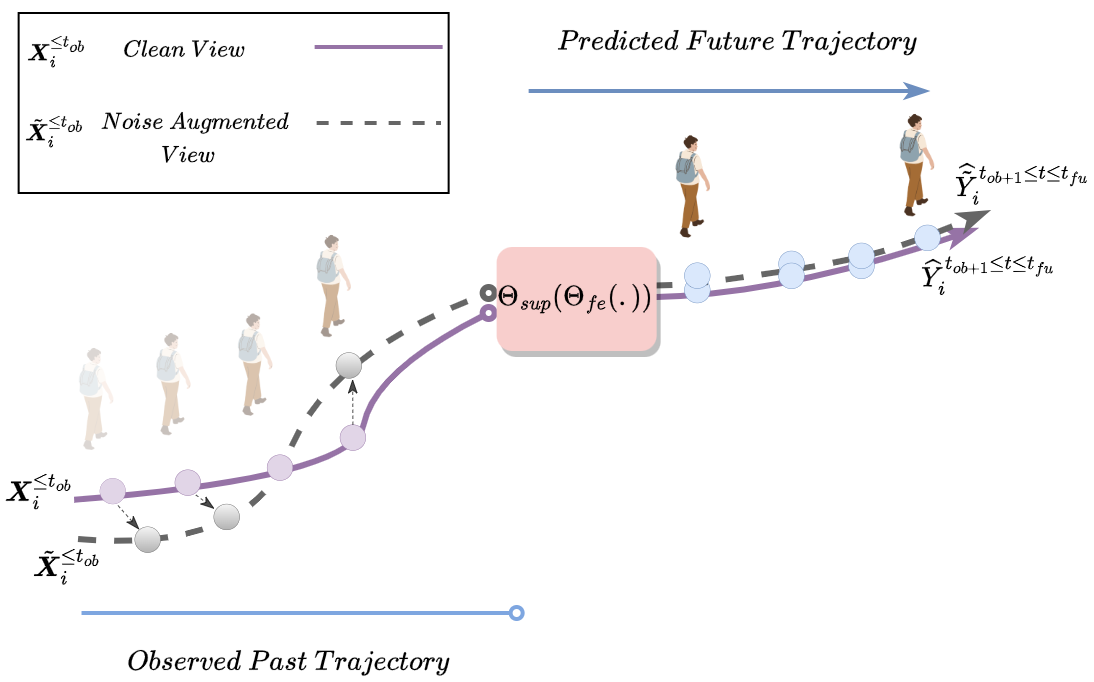}
    \caption{Illustration of the clean (${\boldsymbol{X}}_{i}^{\leq t_{ob}}$) and noise-augmented view (dotted trajectory $\tilde{\boldsymbol{X}}_{i}^{\leq t_{ob}}$), in which the augmentation spatially relocates waypoints with some additive noise within the observed trajectories. The trajectory prediction model ($\Theta_{sup}$) forecasts the trajectories from both of these views.}
    \label{fig2}
\end{figure}

\subsubsection{Clean and Augmented Views}

In our approach, we first generate two different views of past observed trajectories: one characterized as the clean view and the other as a noise-augmented view. The clean view corresponds to the original past trajectory, while the noise-augmented view corresponds to the past trajectory that has been spatially relocated with some additive noise (see Figure~\ref{fig2}).

Given the observed past trajectory $X_{i}^{\leq t_{ob}}$ of agent $i$, the clean view and noise-augmented view are denoted by $X_{i}^{\leq t_{ob}}$ and $\tilde X_{i}^{\leq t_{ob}}$ respectively.
We sample the Gaussian noise ($\mathcal{N}(0, 1)$) and add it to $X_{i}^{\leq t_{ob}}$ to create the noise-augmented view. Specifically, ${\Phi^{\prime}}_{i}^{\leq t_{ob}} \sim \mathcal{N}(0, 1)$ is noise sampled from the standard normal distribution. We control this noise by a parameter ($\omega$) called the noise factor to get the final additive noise ($\Phi_{i}^{\leq t_{ob}}$) as shown in  Equation \ref{eq_noise_addd}. We have also provided an ablation in Section \ref{ab_noisefactor} on choosing the appropriate $\omega$ value. The noise factor controls the spatial relocation of waypoints in the noise-augmented view.

\begin{equation} \label{eq_noise_addd}
    \Phi_{i}^{\leq t_{ob}} = \omega*{\Phi^{\prime}}_{i}^{\leq t_{ob}}
\end{equation}

\begin{equation}
\boldsymbol{\tilde{X}}_{i}^{\leq t_{ob}} = \boldsymbol{X}_{i}^{\leq t_{ob}} + \boldsymbol{\Phi}_{i}^{\leq t_{ob}}
\end{equation}

\begin{equation}
\tilde{\boldsymbol X}_{i}^{\leq t_{ob}} =  \{\boldsymbol{x}_{i}^{t_1},...,\boldsymbol{x}_{i}^{t_{ob}}\} + \{\boldsymbol\sigma^{t_{1}}_{i},...,\boldsymbol\sigma^{t_{ob}}_{i}\} 
\end{equation}

We add additive noise $\boldsymbol\Phi^{\leq t_{ob}}_{i} = \{\boldsymbol\sigma^{t_{1}}_{i},...,\boldsymbol\sigma^{t_{ob}}_{i}\}$ to the past observed trajectory $\boldsymbol X_{i}^{\leq t_{ob}} = \{\boldsymbol{x}_{i}^{t_1},...,\boldsymbol{x}_{i}^{t_{ob}}\}$ of agent $i$ to obtain the noise-augmented view ($\tilde{\boldsymbol X}_{i}^{\leq t_{ob}}$). Here, $\boldsymbol{\sigma}_{i}^{t_{ob}} \in \mathbb{R}^{2}$ represents the 2D Gaussian noise vector for agent $i$ at time step $t_{ob}$.

\subsubsection{Spatial Consistency Module}\label{sec_spatial_cons}

After creating clean and augmented views for agent $i$, we feed them as input to the feature extraction network ($\Theta_{fe}$). The feature extraction network generates features corresponding to both the clean view and the noise-augmented view. The features from the clean view are then input into the trajectory prediction network ($\Theta_{sup}$) to predict the feature trajectory ($\widehat{Y}_{i}$). Similarly, the features from the noise-augmented view are also passed through the trajectory prediction network ($\Theta_{sup}$) to obtain the future trajectory corresponding to the noise-augmented view, as indicated in the equations below.

\begin{equation}
\widehat {Y}_{i}^{t_{ob+1} \leq t \leq t_{fu}} = \Theta_{sup} (\Theta_{fe} (X_{i}^{\leq t_{ob}}))
\end{equation}

\begin{equation}
    \widehat{\tilde{Y}}_{i}^{t_{ob+1} \leq t \leq t_{fu}} = \Theta_{sup} (\Theta_{fe} (\tilde X_{i}^{\leq t_{ob}})
\end{equation}

Here, $\widehat{Y}_{i}$ and $\widehat{\tilde{Y}}_{i}$ denote the future trajectory predictions from the clean and augmented views of the past observed trajectory, respectively. Next, we use the trajectory prediction loss ($\mathcal{L}_{\text{tp}}$) to minimize the gap between the predicted trajectory and the ground truth trajectory. The supervised loss ($L_{\text{sup}}$) is defined as shown in Equation \ref{eq_6}. It is evident from Equation \ref{eq_6} that we are minimizing the gap between ${\widehat Y}_{i}$ and ${Y}_{i}$. Simultaneously, we are also minimizing the gap between ${\widehat{\tilde{Y}}}_{i}$ and ${Y}_{i}$, thus implicitly minimizing the gap between ${\widehat Y}_{i}$ and ${\widehat{\tilde{Y}}}_{i}$. Therefore, $\mathcal{L}_{\text{sup}}$ maintains spatial consistency between the future predictions from clean observed trajectories and the noise-augmented trajectories, bringing predictions from both closer to the ground truth trajectories. This consistency ensures that both views are consistent with each other.

\begin{equation}
\label{eq_6}
\begin{aligned}
\mathcal{L}_{\text{sup}} =
\mathop{\text{}} \frac{1}{N} \sum_{i=1}^N \Big( \mathcal{L}_{\text{tp}}( {\widehat Y}_{i}^{t_{ob+1} \leq t \leq t_{fu}}, { Y}_{i}^{t_{ob+1} \leq t \leq t_{fu}}) + \\
\mathcal{L}_{\text{tp}}(  {\widehat{\tilde{Y}}}^{t_{ob+1} \leq t \leq t_{fu}}, { Y}_{i}^{t_{ob+1} \leq t \leq t_{fu}} ) \Big)
\end{aligned}
\end{equation}

Where $N$ is the number of agents, $Y_i$ is the ground truth future trajectory for agent $i$.

\subsubsection{Noise Prediction Module}\label{sec_noisepred}

The self-supervised noise prediction task involves predicting the noise present in both the clean view (observed past trajectory $X_{i}^{\leq t_{ob}}$) and the noise-augmented view ($\tilde X_{i}^{\leq t_{ob}}$). Specifically, the goal is to estimate the noise value associated with a given observed waypoint.

\begin{equation} \label{eq.noise_from_views}
    \begin{aligned}
        \widehat{\tilde{\Phi}}_{i}^{\leq t_{ob}} = \Theta_{ss} (\Theta_{fe} (\tilde X_{i}^{\leq t_{ob}})) \\
        \widehat{\Phi}_{i}^{\leq t_{ob}} = \Theta_{ss} (\Theta_{fe} ( X_{i}^{\leq t_{ob}}))
    \end{aligned}
\end{equation}

Where $\Theta_{ss}$ represents the parameters of the noise prediction network. $\widehat{\tilde{\Phi}}_{i}^{\leq t_{ob}}$ is the predicted noise in the augmented view for agent $i$, $\widehat{\Phi}_{i}^{\leq t_{ob}}$ is the predicted noise for the clean view of agent $i$. ${{\Phi}}_{i}^{\leq t_{ob}}$ is the ground truth noise (see  Equation \ref{eq_noise_addd}). Please note that the features extracted by $\Theta_{fe}$ are utilized as input to $\Theta_{ss}$ (the noise predicting network) for predicting the noise in the observed past trajectories (clean and noise-augmented views as shown in  Equation \ref{eq.noise_from_views}).

The self supervised prediction loss  $\mathcal{L}_{\text{ss}}$, is given by:

\begin{equation}\label{eq1}
\mathcal{L}_{\text{ss}} = 
\mathop{\text{}} \frac{1}{N} \sum_{i=1}^N \Big( \mathcal{L}_{\text{mse}}( {\widehat{\Phi}}_{i}^{\leq t_{ob}}, 0 ) + \mathcal{L}_{\text{mse}}( \widehat{\tilde{\Phi}}_{i}^{\leq t_{ob}}, {{\Phi}}_{i}^{\leq t_{ob}} )  \Big)
\end{equation}

Here, MSE refers to mean square error. Please note that a value of 0 signifies the absence of noise in the clean view, indicating that no noise is present in the original past observed trajectories ($X_{i}^{\leq t_{ob}}$) of agent $i$.

\subsection{Learning and Evaluation}

The total loss is given as:

\begin{equation}
    \mathcal{L}_{\text{total}} = \mathcal{L}_{\text{sup}} + \lambda\*\mathcal{L}_{\text{ss}}
\end{equation}

\begin{equation} \label{eq:5}
\boldsymbol{\mathrm{\Theta^{\star}_{fe}}}, \boldsymbol{\mathrm{\Theta^{\star}_{sup}}},\boldsymbol{\mathrm{\Theta^{\star}_{ss}}}  =  \mathop{\text{arg min}}_{ \boldsymbol{\mathrm{\Theta_{fe}},\Theta_{sup},\Theta_{ss}}}  \mathcal{L}_{\text{total}}
\end{equation}

Here, $\mathcal{L}_{\text{total}}(\cdot)$ denotes the total loss for training the SSWNP. Additionally, $\lambda$ signifies the contribution of the self-supervised loss in the total loss for training the model using our approach. Given the past observed trajectory, we can predict the future trajectory using Equation~\ref{eq:4} at the test time.

\begin{equation} \label{eq:4}
\begin{aligned}
\widehat {Y}_{i}^{t_{ob+1} \leq t \leq t_{fu}}=\mathrm{\Theta^{\star}_{sup}}(\mathrm{\Theta^{\star}_{fe}}(X_i^{\leq t_{ob}}))
\end{aligned}
\end{equation}

\begin{table}[t]
\caption{Noise factor ($\omega$) and $\lambda$ values used in our experiments.}
\label{tab:exp_implementation}
\resizebox{\columnwidth}{!}{%
\begin{tabular}{c|c|c|c}
\hline
Dataset      & Baselines & $\omega$ & $\lambda$   \\ \hline
NBA          & GroupNet       & $5*10^{-2}$           & $10^{-2}$ \\ \hline
TrajNet      & AutoBot        & $10^{-1}$             & $10^{-1}$ \\ \hline
ETH, UNIV    & Graph-TERN, SSAGCN        & $10^{-2}$             & $10^{-1}$ \\ \hline
ZARA1, ZARA2 & Graph-TERN, SSAGCN         & $10^{-1}$             & $10^{-1}$ \\ \hline
HOTEL        & Graph-TERN, SSAGCN         & $10^{-3}$             & $10^{-1}$ \\ \hline
\end{tabular}%
}
\end{table}

\begin{table}[]
\caption{The minimum Average Displacement Error (minADE) and minimum Final Displacement Error (minFDE) for prediction on the NBA dataset using the \textit{SSWNP} approach. (B) denotes the baseline GroupNet model. RD($\%$) indicates the relative percentage difference compared to the baseline.}
\label{nba_dataset}
\resizebox{\columnwidth}{!}{%
\begin{tabular}{c|cccc}
\hline
\multirow{2}{*}{\textbf{Method}} &
  \multicolumn{4}{c}{\textbf{Time}} \\ \cline{2-5} 
 &
  \textbf{1.0s} &
  \textbf{2.0}s &
  \textbf{3.0s} &
  \textbf{4.0s} \\ \hline
\begin{tabular}[c]{@{}c@{}}SLSTM   \cite{alahi2016social}\end{tabular} &
  \multicolumn{1}{c|}{0.45/0.67} &
  \multicolumn{1}{c|}{0.88/1.53} &
  \multicolumn{1}{c|}{1.33/2.38} &
  1.79/3.16 \\ \hline
\begin{tabular}[c]{@{}c@{}}SGAN  \cite{gupta2018social}\end{tabular} &
  \multicolumn{1}{c|}{0.46/0.65} &
  \multicolumn{1}{c|}{0.85/1.36} &
  \multicolumn{1}{c|}{1.24/1.98} &
  1.62/2.51 \\ \hline
\begin{tabular}[c]{@{}c@{}}SSTGCNN \cite{mohamed2020social}\end{tabular} &
  \multicolumn{1}{c|}{0.36/0.50} &
  \multicolumn{1}{c|}{0.75/0.99} &
  \multicolumn{1}{c|}{1.15/1.79} &
  1.59/2.37 \\ \hline
\begin{tabular}[c]{@{}c@{}}STGAT \cite{huang2019stgat}\end{tabular} &
  \multicolumn{1}{c|}{0.38/0.55} &
  \multicolumn{1}{c|}{0.73/1.18} &
  \multicolumn{1}{c|}{1.07/1.74} &
  1.41/2.22 \\ \hline
\begin{tabular}[c]{@{}c@{}}NRI \cite{pmlr-v80-kipf18a}\end{tabular} &
  \multicolumn{1}{c|}{0.45/0.64} &
  \multicolumn{1}{c|}{0.84/1.44} &
  \multicolumn{1}{c|}{1.24/2.18} &
  1.62/2.84 \\ \hline
\begin{tabular}[c]{@{}c@{}}STAR \cite{yu2020spatio}\end{tabular} &
  \multicolumn{1}{c|}{0.43/0.65} &
  \multicolumn{1}{c|}{0.77/1.28} &
  \multicolumn{1}{c|}{1.00/1.55} &
  1.26/2.04 \\ \hline
\begin{tabular}[c]{@{}c@{}}PECNet \cite {mangalam2020not}\end{tabular} &
  \multicolumn{1}{c|}{0.51/0.76} &
  \multicolumn{1}{c|}{0.96/1.69} &
  \multicolumn{1}{c|}{1.41/2.52} &
  1.83/3.41 \\ \hline
\begin{tabular}[c]{@{}c@{}}NMMP \cite{hu2020collaborative}\end{tabular} &
  \multicolumn{1}{c|}{0.38/0.54} &
  \multicolumn{1}{c|}{0.70/1.11} &
  \multicolumn{1}{c|}{1.01/1.61} &
  1.33/2.05 \\ \hline
\begin{tabular}[c]{@{}c@{}}DynGroupNet \cite{xu2023dynamic}\end{tabular} &
  \multicolumn{1}{c|}{0.19/0.28} &
  \multicolumn{1}{c|}{0.40/0.61} &
  \multicolumn{1}{c|}{0.65/0.90} &
  0.89/1.13 \\ \hline

\begin{tabular}[c]{@{}c@{}}Stimulus \cite{sun2023stimulus}\end{tabular} &
  \multicolumn{1}{c|}{-} &
  \multicolumn{1}{c|}{-} &
  \multicolumn{1}{c|}{-} &
  1.08/1.12 \\ \hline
\begin{tabular}[c]{@{}c@{}}TDGCN \cite{wang2023trajectory}\end{tabular} &
  \multicolumn{1}{c|}{0.30/0.45} &
  \multicolumn{1}{c|}{0.53/0.82} &
  \multicolumn{1}{c|}{0.80/117} &
  1.06/153 \\ \hline
\begin{tabular}[c]{@{}c@{}}MERA \cite{sun2023modality}\end{tabular} &
  \multicolumn{1}{c|}{-} &
  \multicolumn{1}{c|}{-} &
  \multicolumn{1}{c|}{-} &
  1.17/2.21 \\ \hline
\begin{tabular}[c]{@{}c@{}}DISTL \cite{cao2023discovering}\end{tabular} &
  \multicolumn{1}{c|}{0.30/0.40} &
  \multicolumn{1}{c|}{0.58/0.88} &
  \multicolumn{1}{c|}{0.87/1.31} &
  1.13/1.60 \\ \hline
\begin{tabular}[c]{@{}c@{}}GroupNet (B) \cite{Xu_2022_CVPR}\end{tabular} &
  \multicolumn{1}{c|}{0.34/0.48} &
  \multicolumn{1}{c|}{0.62/0.95} &
  \multicolumn{1}{c|}{0.87/1.31} &
  1.13/1.69 \\ \hline
\begin{tabular}[c]{@{}c@{}}Our (B)+SSWNP\end{tabular} &
  \multicolumn{1}{c|}{\textbf{0.23/0.31}} &
  \multicolumn{1}{c|}{\textbf{0.45/0.63}} &
  \multicolumn{1}{c|}{\textbf{0.67/0.92}} &
  \textbf{0.90/1.14} \\ \hline
\begin{tabular}[c]{@{}c@{}}RD($\%$) ADE/FDE\end{tabular} &
  \multicolumn{1}{l|}{\textbf{38.6/43.0}} &
  \multicolumn{1}{l|}{\textbf{31.8/40.5}} &
  \multicolumn{1}{l|}{\textbf{26.0/35.0}} &
  \multicolumn{1}{l}{\textbf{22.7/38.9}} \\ \hline

\end{tabular}%
}
\end{table}

\begin{table}[]
\centering
\caption{Quantitative results using the \textit{SSWNP} approach on the TrajNet++ dataset during training. (B) stands for the baseline model. RD($\%$) indicates the relative percentage difference from the baseline.
}
\label{trajnet_dataset}
\resizebox{\columnwidth}{!}{%
\begin{tabular}{c|c|c}
\hline
\textbf{Model} &
  \textbf{\begin{tabular}[c]{@{}l@{}}Scene-level \\ Min ADE ($\downarrow$) \end{tabular}} &
  \textbf{\begin{tabular}[c]{@{}l@{}}Scene-level \\ Min FDE ($\downarrow$) \end{tabular}} \\ \hline
\multicolumn{1}{c|}{Social gan \cite{gupta2018social}} & \multicolumn{1}{c|}{0.57}          & 1.24          \\ \hline
\multicolumn{1}{c|}{Social attention \cite{vemula2018social}} & \multicolumn{1}{c|}{0.56}          & 1.21          \\ \hline
\multicolumn{1}{c|}{Social-bigat \cite{kosaraju2019social} } & \multicolumn{1}{c|}{0.56}          & 1.22         \\ \hline
\multicolumn{1}{c|}{trajectron \cite{ivanovic2019trajectron} } & \multicolumn{1}{c|}{0.60}          & 1.28          \\ \hline
\multicolumn{1}{c|}{AIN \cite{zhu2020robust}} & \multicolumn{1}{c|}{0.620}          & 1.240         \\ \hline
\multicolumn{1}{c|}{PecNet \cite{mangalam2020not} } & \multicolumn{1}{c|}{0.570}          & 1.180         \\ \hline
\multicolumn{1}{c|}{AMENet \cite{cheng2021amenet}  } & \multicolumn{1}{c|}{0.620}          & 1.300         \\ \hline
\multicolumn{1}{c|}{socially-aware \cite{saadatnejad2022socially} } & \multicolumn{1}{c|}{0.60}          & 1.28          \\ \hline
\multicolumn{1}{c|}{Linear Extrapolation \cite{girgis2022latent} } & \multicolumn{1}{c|}{0.409}          & 0.897          \\ \hline
\multicolumn{1}{c|}{AntiSocial \cite{girgis2022latent} }   & \multicolumn{1}{c|}{0.316}          & 0.632          \\ \hline
\multicolumn{1}{c|}{Ego \cite{girgis2022latent} }          & \multicolumn{1}{c|}{0.214}          & 0.431          \\ \hline
\multicolumn{1}{c|}{AutoBot (B) \cite{girgis2022latent} }              & \multicolumn{1}{c|}{0.128}          & 0.234          \\ \hline
\multicolumn{1}{c|}{Our (B)+SSWNP}                  & \multicolumn{1}{c|}{\textbf{0.091}} & \textbf{0.162} \\ \hline
\multicolumn{1}{c|}{\begin{tabular}[c]{@{}c@{}} RD($\%$)\end{tabular}} &
  \multicolumn{1}{c|}{\textbf{33.8}} &
  \textbf{36.4} \\ \hline
\end{tabular}}
\end{table}

\section{Experiments}
In this Section, we present the quantitative and qualitative results of our approach. Additionally, we have conducted several ablation studies.
\subsection{Experimental Details}

\subsubsection{Dataset} 

We evaluate the performance of SSWNP on three trajectory datasets: NBA \cite{zhan2018generating}, TrajNet++ \cite{Kothari2020HumanTF}, and ETH-UCY \cite{pellegrini2009you, lerner2007crowds}. The NBA Sports VU Dataset includes player trajectory data from all ten players in live NBA games, where teammates heavily influence player motions. In this assessment, we predict the following ten timestamps (4.0 seconds) using the five timestamps that occurred before them, spanning 2.0 seconds of past data. The key objective of TrajNet++ is to highlight significant agent-agent interactions across a scenario. Specifically, we evaluate the model for the subsequent 12 timestamps based on the agents' last nine timestamps. ETH-UCY is a composite of two datasets featuring smooth trajectories and straightforward agent interactions. The ETH dataset includes two scenarios, ETH and HOTEL, totaling 750 pedestrians. On the other hand, UNIV, ZARA1, and ZARA2 scenarios, totaling 786 pedestrians, are included in the UCY dataset. These scenes encompass various settings, including roads, intersections, and open areas. The world-coordinate sequence comprises trajectories covering eight time steps, or 3.2 seconds. We aim to forecast the next 12 time steps, so our predictions will cover 4.8 seconds in total.

\subsubsection{Evaluation Metric} We use standard evaluation metrics such as Average Displacement Error (ADE) and Final Displacement Error (FDE) for trajectory prediction evaluation. ADE represents the average L2 distance between predicted and ground truth trajectories across all time steps. In contrast, FDE quantifies the L2 distance at the last time step or final endpoint.

\subsubsection{Implementation Details}  

\begin{figure}
    \centering
    \includegraphics[scale=.15]{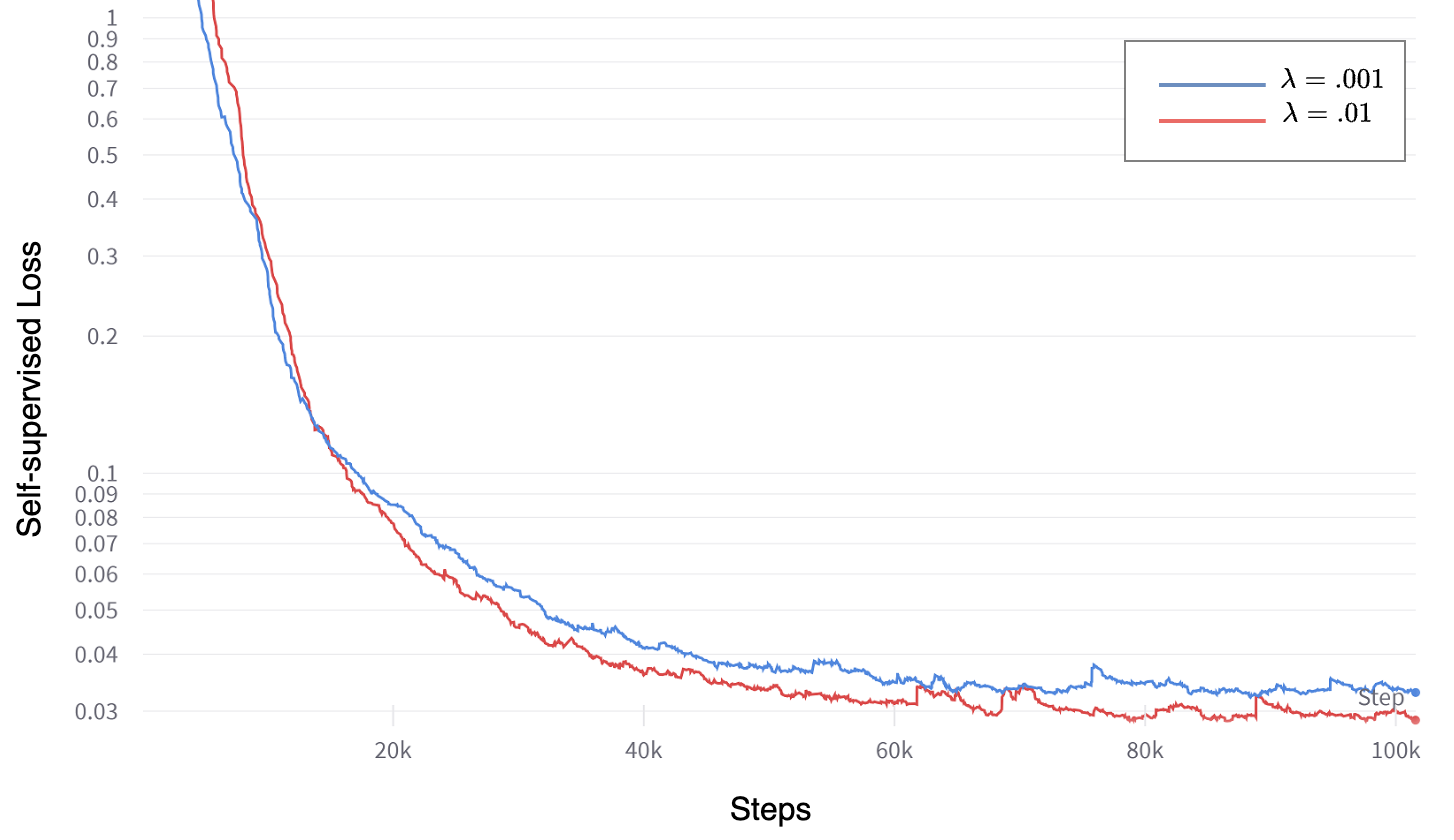}
    \caption{Illustration of the self-supervised loss ($\mathcal{L}_{\text{ss}}$) plot for \textit{GroupNet+SSWNP}, indicating a decrease in loss value over the training steps on the NBA dataset. The optimal hyperparameter value for $\lambda$ is chosen to be 0.01 (shown in red), suggesting improved learning facilitated by the noise prediction network.}
    \label{plot_lambda}
\end{figure}

To ensure a fair comparison with the methods under consideration, we maintained their default configurations, including the trajectory sequence length and timestamps used as model input. We selected the $\lambda$ value based on the convergence of the self-supervised loss, and one such plot for GroupNet is shown in Figure~\ref{plot_lambda}. We considered Gaussian noise with a mean of 0 and a standard deviation of 1 for sampling noise. The noise factor ($\omega$) is used to generate a noise-augmented view, while $\lambda$  defines the contribution of the self-supervised loss to the total training loss. The values for the noise factor ($\omega$) and $\lambda$ used in our experimentation during the training of the model are provided in Table~\ref{tab:exp_implementation}. Section~\ref{ab_noisefactor} provides insight into the choice of the noise factor.

\subsubsection{Baseline Models} We assess our approach by testing it on four distinct models: a Variational Autoencoder-based model (GroupNet \cite{Xu_2022_CVPR}), Transformer-based model (AutoBot \cite{girgis2022latent}), Graph-based (SSAGCN \cite{lv2023ssagcn}) and Goal-based model (Graph-TERN \cite{bae2023set}). GroupNet excels at capturing interactions among agents, allowing it to anticipate socially plausible trajectories using relational reasoning. When combined with a Conditional Variational Autoencoder (CVAE), GroupNet can learn complex social variables for better trajectory prediction. AutoBot is an encoder-decoder architecture utilizing transformers to construct multi-agent trajectories consistent with the scene. In this architecture, the encoder employs alternating temporal and social multi-head self-attention mechanisms to facilitate learning across time and social dimensions. The SSAGCN models the degree of influence among pedestrians using a spatial-temporal graph and forecasts trajectories that align with both social and physical feasibility. Graph-TERN captures social and temporal relationships through a pedestrian graph and then employs control point prediction to refine trajectories. Graph-TERN also overcomes accumulated errors through control points and intermediate destinations.

\subsubsection{Architecture Details}
The SSWNP architecture comprises three primary components, as illustrated in Figure~\ref{first_fig}: the feature extractor network, the trajectory prediction network, and the noise prediction network. The feature extraction network ($\Theta_{fe}$) generates features for both clean and noise-augmented views. For GroupNet, $\Theta_{fe}$ represents the encoder of CVAE; for Autobot, it is the encoder of the transformer; for GraphTern, it is the multi relational graph convolutional network; and for SSAGCN, it is convolutional neural network. The trajectory prediction network ($\Theta_{sup}$) predicts the future trajectory. For GroupNet, the trajectory prediction network is the decoder of CVAE. For AutoBot, the trajectory prediction network is the decoder of the transformer. For GraphTern, it is the graph convolutional network. For SSAGCN, it is the temporal convolutional neural network. The noise prediction network ($\Theta_{ss}$) predicts the noise present in the observed past trajectory. The noise prediction network is a multilayer perceptron (MLP). The input layer dimension of MLP is the dimension of output produced by the feature extraction network. The output layer dimension of MLP is the dimension of past observed trajectory. There are two hidden layers in MLP, with 128 and 64 nodes in the first and second hidden layers, respectively.

\begin{table*}[]
\caption{Minimum ADE ($\downarrow$) / Minimum FDE ($\downarrow$) for trajectory prediction on the ETH-UCY dataset utilizing the \textit{SSWNP} technique during training. (B1) and (B2) denote the first and second baseline models. RD1 ($\%$) and RD2 ($\%$) indicate the relative percentage difference compared to the baseline B1 and B2, respectively.}
\label{table:eth-ucy}
\centering
\begin{tabular}{c|ccccc|c}
\hline
\textbf{Method} &
  \textbf{ETH} &
  \textbf{HOTEL} &
  \textbf{UNIV} &
  \textbf{ZARA1} &
  \textbf{ZARA2} &
  \textbf{AVG} \\ \hline
SGAN \cite{gupta2018social} &
  \multicolumn{1}{c|}{0.87/1.62} &
  \multicolumn{1}{c|}{0.67/1.37} &
  \multicolumn{1}{c|}{0.76/1.52} &
  \multicolumn{1}{c|}{0.35/0.68} &
  0.42/0.84 &
  0.61/1.21 \\ \hline
Sophie \cite{sophie19} &
  \multicolumn{1}{c|}{0.70/1.43} &
  \multicolumn{1}{c|}{0.76/1.67} &
  \multicolumn{1}{c|}{0.54/1.24} &
  \multicolumn{1}{c|}{0.30/0.63} &
  0.38/0.78 &
  0.54/1.15 \\ \hline
STGAT \cite{sekhon2021scan} &
  \multicolumn{1}{c|}{0.56/1.10} &
  \multicolumn{1}{c|}{0.27/0.50} &
  \multicolumn{1}{c|}{0.32/0.66} &
  \multicolumn{1}{c|}{0.21/0.42} &
  0.20/0.40 &
  0.31/0.62 \\ \hline
Social-BiGAT \cite{kosaraju2019social} &
  \multicolumn{1}{c|}{0.69/1.29} &
  \multicolumn{1}{c|}{0.49/1.01} &
  \multicolumn{1}{c|}{0.55/1.32} &
  \multicolumn{1}{c|}{0.30/0.62} &
  0.36/0.75 &
  0.48/1.00 \\ \hline
NMMP \cite{hu2020collaborative} &
  \multicolumn{1}{c|}{0.62/1.08} &
  \multicolumn{1}{c|}{0.33/0.63} &
  \multicolumn{1}{c|}{0.52/1.11} &
  \multicolumn{1}{c|}{0.32/0.66} &
  0.29/0.61 &
  0.41/0.82 \\ \hline
Social-STGCNN \cite{mohamed2020social} &
  \multicolumn{1}{c|}{0.64/1.11} &
  \multicolumn{1}{c|}{0.49/0.85} &
  \multicolumn{1}{c|}{0.44/0.79} &
  \multicolumn{1}{c|}{0.34/0.53} &
  0.30/0.48 &
  0.44/0.75 \\ \hline
CARPE \cite{mendieta2021carpe} &
  \multicolumn{1}{c|}{0.80/1.4} &
  \multicolumn{1}{c|}{0.52/1.00} &
  \multicolumn{1}{c|}{0.61/1.23} &
  \multicolumn{1}{c|}{0.42/0.84} &
  0.34/0.74 &
  0.46/0.89 \\ \hline
PecNet \cite {mangalam2020not} &
  \multicolumn{1}{c|}{0.54/0.87} &
  \multicolumn{1}{c|}{0.18/0.24} &
  \multicolumn{1}{c|}{0.35/0.60} &
  \multicolumn{1}{c|}{0.22/0.39} &
  0.17/0.30 &
  0.29/0.48 \\ \hline
Trajectron++ \cite{salzmann2020trajectron++} &
  \multicolumn{1}{c|}{0.43/0.86} &
  \multicolumn{1}{c|}{0.12/0.19} &
  \multicolumn{1}{c|}{0.22/0.43} &
  \multicolumn{1}{c|}{0.17/0.32} &
  0.12/0.25 &
  0.21/0.41 \\ \hline
GTPPO \cite{yang2021novel} &
  \multicolumn{1}{c|}{0.63/0.98} &
  \multicolumn{1}{c|}{0.19/0.30} &
  \multicolumn{1}{c|}{0.35/0.60} &
  \multicolumn{1}{c|}{0.20/0.32} &
  0.18/0.31 &
  0.31/0.50 \\ \hline
SGCN \cite{shi2021sgcn} &
  \multicolumn{1}{c|}{0.52/1.03} &
  \multicolumn{1}{c|}{0.32/0.55} &
  \multicolumn{1}{c|}{0.37/0.70} &
  \multicolumn{1}{c|}{0.29/0.53} &
  0.25/0.45 &
  0.37/0.65 \\ \hline
Introvert \cite{shafiee2021introvert} &
  \multicolumn{1}{c|}{0.42/0.70} &
  \multicolumn{1}{c|}{0.11/0.17} &
  \multicolumn{1}{c|}{0.20/0.32} &
  \multicolumn{1}{c|}{0.16/0.27} &
  0.16/0.25 &
  0.21/0.34 \\ \hline
LB-EBM \cite{pang2021trajectory} &
  \multicolumn{1}{c|}{0.30/0.52} &
  \multicolumn{1}{c|}{0.13/0.20} &
  \multicolumn{1}{c|}{0.27/0.52} &
  \multicolumn{1}{c|}{0.20/0.37} &
  0.15/0.29 &
  0.21/0.38 \\ \hline
\begin{tabular}[c]{@{}c@{}}GroupNet  \cite{Xu_2022_CVPR}\end{tabular} &
  \multicolumn{1}{c|}{0.46/0.73} &
  \multicolumn{1}{c|}{0.15/0.25} &
  \multicolumn{1}{c|}{0.26/0.49} &
  \multicolumn{1}{c|}{0.21/0.39} &
  \multicolumn{1}{c|}{0.17/0.33} &
  0.25/0.44 \\ \hline
\begin{tabular}[c]{@{}c@{}}DynGroupNet \cite{xu2023dynamic}\end{tabular} &
  \multicolumn{1}{c|}{0.42/0.66} &
  \multicolumn{1}{c|}{0.13/0.20} &
  \multicolumn{1}{c|}{0.24/0.44} &
  \multicolumn{1}{c|}{0.19/0.34} &
  \multicolumn{1}{c|}{0.15/0.28} &
  0.23/0.38 \\ \hline
\begin{tabular}[c]{@{}c@{}}TDGCN \cite{wang2023trajectory}\end{tabular} &
  \multicolumn{1}{c|}{0.51/0.68} &
  \multicolumn{1}{c|}{0.25/0.44} &
  \multicolumn{1}{c|}{0.30/0.50} &
  \multicolumn{1}{c|}{0.24/0.42} &
  \multicolumn{1}{c|}{0.16/0.27} &
  0.29/0.46 \\ \hline
\begin{tabular}[c]{@{}c@{}}MERA \cite{sun2023modality}\end{tabular} &
  \multicolumn{1}{c|}{0.26/0.50} &
  \multicolumn{1}{c|}{0.11/0.19} &
  \multicolumn{1}{c|}{0.25/0.53} &
  \multicolumn{1}{c|}{0.19/0.40} &
  \multicolumn{1}{c|}{0.15/0.31} &
  0.19/0.39 \\ \hline
\begin{tabular}[c]{@{}c@{}}RMB \cite{shi2023representing}\end{tabular} &
  \multicolumn{1}{c|}{0.29/0.49} &
  \multicolumn{1}{c|}{0.12/0.18} &
  \multicolumn{1}{c|}{0.29/0.51} &
  \multicolumn{1}{c|}{0.20/0.36} &
  \multicolumn{1}{c|}{0.15/0.27} &
  0.21/0.36 \\ \hline
\begin{tabular}[c]{@{}c@{}}VIKT  \cite{zhong2023visual}\end{tabular} &
  \multicolumn{1}{c|}{0.30/0.51} &
  \multicolumn{1}{c|}{0.13/0.25} &
  \multicolumn{1}{c|}{0.23/0.51} &
  \multicolumn{1}{c|}{0.21/0.44} &
  \multicolumn{1}{c|}{0.14/0.30} &
  0.20/0.40 \\ \hline

\begin{tabular}[c]{@{}c@{}}MSN \cite{wong2023msn}\end{tabular} &
  \multicolumn{1}{c|}{0.27/0.41} &
  \multicolumn{1}{c|}{0.11/0.17} &
  \multicolumn{1}{c|}{0.28/0.48} &
  \multicolumn{1}{c|}{0.22/0.36} &
  \multicolumn{1}{c|}{0.18/0.29} &
  0.21/0.34 \\ \hline
\begin{tabular}[c]{@{}c@{}}LSSTA \cite{yang2023long}\end{tabular} &
  \multicolumn{1}{c|}{0.30/0.52} &
  \multicolumn{1}{c|}{0.12/0.20} &
  \multicolumn{1}{c|}{0.28/0.55} &
  \multicolumn{1}{c|}{0.20/0.40} &
  \multicolumn{1}{c|}{0.16/0.32} &
  0.21/0.40 \\ \hline
  \begin{tabular}[c]{@{}c@{}}RCPN \cite{zhu2023reciprocal}\end{tabular} &
  \multicolumn{1}{c|}{0.48/0.86} &
  \multicolumn{1}{c|}{0.38/0.68} &
  \multicolumn{1}{c|}{0.31/0.58} &
  \multicolumn{1}{c|}{0.25/0.44} &
  \multicolumn{1}{c|}{0.23/0.35} &
  0.33/0.58 \\ \hline
  \begin{tabular}[c]{@{}c@{}}STS LSTM \cite{zhang2023spatial}\end{tabular} &
  \multicolumn{1}{c|}{0.46/0.81} &
  \multicolumn{1}{c|}{ 0.20/0.29} &
  \multicolumn{1}{c|}{0.38/0.70} &
  \multicolumn{1}{c|}{0.30/0.57} &
  \multicolumn{1}{c|}{0.24/0.48} &
  0.32/0.57 \\ \hline
\begin{tabular}[c]{@{}c@{}}SIM  \cite{li2023synchronous}\end{tabular} &
  \multicolumn{1}{c|}{0.32/0.53} &
  \multicolumn{1}{c|}{0.32/0.53} &
  \multicolumn{1}{c|}{0.16/0.34} &
  \multicolumn{1}{c|}{0.12/0.25} &
  \multicolumn{1}{c|}{0.09/0.18} &
  0.16/0.29 \\ \hline  
\begin{tabular}[c]{@{}c@{}}SRGAT  \cite{chen2023goal}\end{tabular} &
  \multicolumn{1}{c|}{0.25/0.38} &
  \multicolumn{1}{c|}{0.10/0.15} &
  \multicolumn{1}{c|}{0.21/0.38} &
  \multicolumn{1}{c|}{0.16/0.28} &
  \multicolumn{1}{c|}{0.12/0.21} &
  0.17/0.28 \\ \hline 
\begin{tabular}[c]{@{}c@{}}VNAGT  \cite{chen2023vnagt}\end{tabular} &
  \multicolumn{1}{c|}{0.52/0.88} &
  \multicolumn{1}{c|}{0.16/0.25} &
  \multicolumn{1}{c|}{0.27/0.51} &
  \multicolumn{1}{c|}{0.23/0.44} &
  \multicolumn{1}{c|}{0.18/0.33} &
  0.27/0.48 \\ \hline   
\begin{tabular}[c]{@{}c@{}}MetaTraj w/MemoNet \cite{shi2023metatraj}\end{tabular} &
  \multicolumn{1}{c|}{0.38/0.59} &
  \multicolumn{1}{c|}{0.11/0.16} &
  \multicolumn{1}{c|}{0.22/0.41} &
  \multicolumn{1}{c|}{0.18/0.30} &
  \multicolumn{1}{c|}{0.13/0.26} &
  0.20/0.34 \\ \hline  
\begin{tabular}[c]{@{}c@{}}SOCIAL SAGAN \cite{yang2023social}\end{tabular} &
  \multicolumn{1}{c|}{0.65/1.19} &
  \multicolumn{1}{c|}{0.36/0.70} &
  \multicolumn{1}{c|}{0.54/1.14} &
  \multicolumn{1}{c|}{0.33/0.66} &
  \multicolumn{1}{c|}{0.29/0.61} &
  0.43/0.86 \\ \hline   

\begin{tabular}[c]{@{}c@{}}Graph-TERN (B1) \cite{bae2023set}\end{tabular} &
  \multicolumn{1}{c|}{0.42/0.58} &
  \multicolumn{1}{c|}{0.14/0.23} &
  \multicolumn{1}{c|}{0.26/0.45} &
  \multicolumn{1}{c|}{0.21/0.37} &
  \multicolumn{1}{c|}{0.17/0.29} &
  0.24/0.38 \\ \hline
\begin{tabular}[c]{@{}c@{}}Our (B1)+SSWNP\end{tabular} &
  \multicolumn{1}{c|}{\textbf{0.38/0.48}} &
  \multicolumn{1}{c|}{\textbf{0.14/0.23}} &
  \multicolumn{1}{c|}{\textbf{0.24/0.40}} &
  \multicolumn{1}{c|}{\textbf{0.19/0.32}} &
  \multicolumn{1}{c|}{\textbf{0.15/0.25}} &
  \textbf{0.22/0.33} \\ \hline
\begin{tabular}[c]{@{}c@{}}RD1($\%$) ADE/FDE \end{tabular} &
  \multicolumn{1}{c|}{-} &
  \multicolumn{1}{c|}{-} &
  \multicolumn{1}{c|}{-} &
  \multicolumn{1}{c|}{-} &
  \multicolumn{1}{c|}{-} &
  \textbf{8.60/14.00} \\ \hline  
\begin{tabular}[c]{@{}c@{}}SSAGCN (B2) \cite{lv2023ssagcn}\end{tabular} &
  \multicolumn{1}{c|}{0.21/0.38} &
  \multicolumn{1}{c|}{0.11/0.19} &
  \multicolumn{1}{c|}{0.14/0.25} &
  \multicolumn{1}{c|}{0.12/0.22} &
  \multicolumn{1}{c|}{0.09/0.15} &
  0.13/0.24 \\ \hline
\begin{tabular}[c]{@{}c@{}}Our (B2)+SSWNP\end{tabular} &
  \multicolumn{1}{c|}{0.21/0.38} &
  \multicolumn{1}{c|}{\textbf{0.078/0.104}} &
  \multicolumn{1}{c|}{\textbf{0.10/0.17}} &
  \multicolumn{1}{c|}{\textbf{0.11/0.19}} &
  \multicolumn{1}{c|}{\textbf{0.079/0.118}} &
  \textbf{0.11/0.19} \\ \hline
\begin{tabular}[c]{@{}c@{}}RD2($\%$) ADE/FDE \end{tabular} &
  \multicolumn{1}{c|}{-} &
  \multicolumn{1}{c|}{-} &
  \multicolumn{1}{c|}{-} &
  \multicolumn{1}{c|}{-} &
  \multicolumn{1}{c|}{-} &
  \textbf{16.60/23.20} \\ \hline  
\end{tabular}%

\end{table*}

\begin{table*}[]
\caption{Results for the \textit{GroupNet+SSWNP} model on the NBA dataset. (B) indicates the baseline GroupNet model. (B+SC) denotes the baseline with the spatial consistency module. (B+SC+NP) denotes the baseline with both the spatial consistency module and the noise prediction module. RD($\%$) refers to the relative percent difference with respect to the baseline.}
\label{tab:self_task}
\centering
\begin{tabular}{c|cccc|cccc|c}
\hline
\multirow{2}{*}{\textbf{Method}} &
  \multicolumn{4}{c|}{\textbf{ADE}} &
  \multicolumn{4}{c|}{\textbf{FDE}} &
  \multirow{2}{*}{ADE/FDE RD(\%)} \\ \cline{2-9}
 &
  \multicolumn{1}{c|}{1.0s} &
  \multicolumn{1}{c|}{2.0s} &
  \multicolumn{1}{c|}{3.0s} &
  4.0s &
  \multicolumn{1}{c|}{1.0s} &
  \multicolumn{1}{c|}{2.0s} &
  \multicolumn{1}{c|}{3.0s} &
  4.0s &
   \\ \hline
$B$ &
  \multicolumn{1}{c|}{0.34} &
  \multicolumn{1}{c|}{0.62} &
  \multicolumn{1}{c|}{0.87} &
  1.13 &
  \multicolumn{1}{c|}{0.48} &
  \multicolumn{1}{c|}{0.95} &
  \multicolumn{1}{c|}{1.31} &
  1.69 &
  - \\ \hline
$B+SC$ &
  \multicolumn{1}{c|}{ 0.283} &
  \multicolumn{1}{c|}{ 0.521} &
  \multicolumn{1}{c|}{0.771} &
  1.018 &
  \multicolumn{1}{c|}{0.370} &
  \multicolumn{1}{c|}{0.754} &
  \multicolumn{1}{c|}{1.101} &
  1.362 &
  10.42/21.49 \\ \hline
$B+SC+NP$ &
  \multicolumn{1}{c|}{\textbf{0.239}} &
  \multicolumn{1}{c|}{\textbf{0.452}} &
  \multicolumn{1}{c|}{\textbf{0.678}} &
  \textbf{0.903} &
  \multicolumn{1}{c|}{\textbf{0.311}} &
  \multicolumn{1}{c|}{\textbf{0.637}} &
  \multicolumn{1}{c|}{\textbf{0.920}} &
  \textbf{1.147} &
  \textbf{22.33/38.28}\\ \hline
\end{tabular}%
\end{table*}

\subsection{Quantitative Results}

\subsubsection{Evaluation on the NBA Dataset}
On the NBA dataset, based on past trajectories from the last five timestamps (2.0 seconds), we forecast future trajectories for ten timestamps (4.0 seconds ahead). Table~\ref{nba_dataset} summarizes the results of an evaluation involving several methods. Our findings demonstrate a significant outperformance of our approach compared to others. Notably, at 4.0 seconds, the minimum Average Displacement Error (minADE) and minimum Final Displacement Error (minFDE) decrease to 0.90 and 1.14, respectively (with a relative improvement of 22.7\% and 38.9\% in ADE/FDE) compared to the baseline GroupNet \cite{Xu_2022_CVPR}.

\subsubsection{Evaluation on the TRAJNET++ Dataset}
On the TRAJNET++ Dataset, leveraging data from the preceding nine timestamps, we forecast the subsequent 12 timestamps for each agent. The integration of \textit{SSWNP} into the AutoBot baseline showcased enhanced performance compared to the baseline, as illustrated in Table~\ref{trajnet_dataset}. Notably, there is a substantial 33.8\% improvement in ADE and a 36.4\% improvement in FDE values when compared to the baseline AutoBot model.

\subsubsection{Evaluation on the ETH-UCY Dataset}
Table~\ref{table:eth-ucy} presents the ADE and FDE values for various methods. Our approach demonstrates enhanced prediction accuracy compared to others. Specifically, with the inclusion of SSWNP, we achieved a relative percentage gain of 8.60/14.00\% with respect to baseline one (B1) and 16.60/23.20\% with respect to baseline two (B2) in ADE/FDE.

\begin{figure*}
    \centering
    \includegraphics[scale=.18]{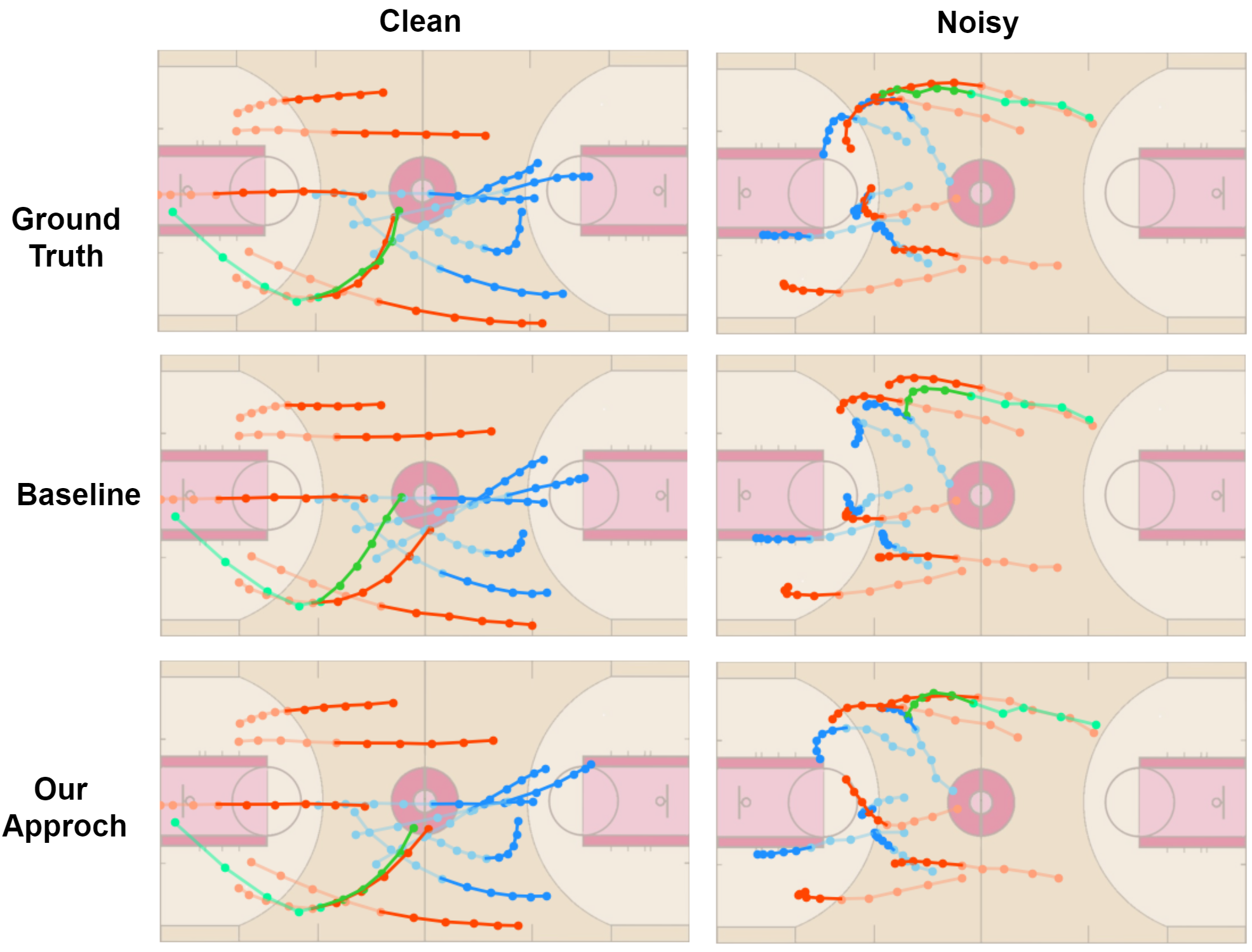}
    \caption{Visual representation of results on the NBA dataset. Trajectories of ten players from each team (\textcolor{cyan}{cyan} and \textcolor{red}{red}) are depicted alongside GroupNet \cite{Xu_2022_CVPR} and the ground truth for comparison. Past trajectories are represented in a lighter color, while predicted waypoints are shown in a solid color. The \textcolor{green}{green} color represents the ball trajectory. The first and second columns display the model's predictions for the next ten timestamps in both clean and noisy environments.}
    \label{nba_vis}
\end{figure*}

\begin{figure}[]
    \centering
    \includegraphics[scale=.1]{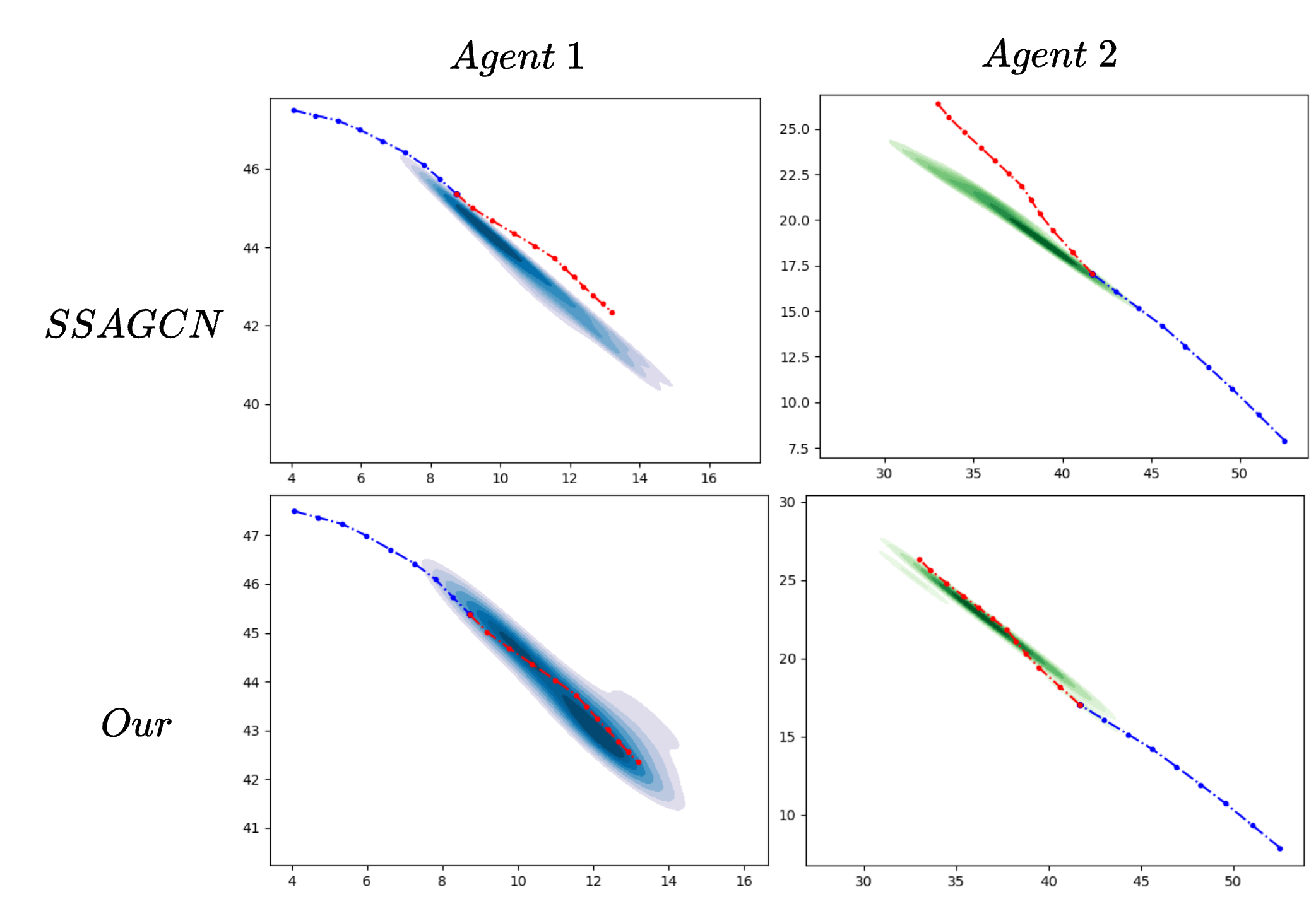}
\caption{Illustration of temporal density estimations of the agent for the ETH/UCY datasets using SSAGCN \cite{lv2023ssagcn} and our approach. The color density (\textcolor{blue}{blue} for Agent 1 and \textcolor{green}{green} for Agent 2) depicts the forecasted distribution of future trajectories, with the blue dotted line representing the historical trajectory (8 timestamps) and the red dotted line corresponding to the actual ground truth (12 timestamps).}
    \label{plot_eth2}
\end{figure}

\subsection{Qualitative Results on NBA Dataset}

We further evaluated the capabilities of our approach through qualitative results. Figure~\ref{nba_vis} illustrates the predictions of our SSWNP and GroupNet in both clean and noisy settings on the NBA SportVU dataset. It is evident from Figure~\ref{nba_vis} that our approach performs better in both clean and noisy environments. The results demonstrate that our approach consistently produces more accurate predictions than the baseline.

\subsection{Qualitative Results on ETH-UCY Dataset}

We have provided visualizations of predicted density on the ETH/UCY datasets as shown in Figure~\ref{plot_eth2}. Our approach effectively captures the agent's future distribution by accurately predicting the future density represented by the blue color (Agent 1) and green color (Agent 2). In contrast to SSAGCN, which predicts the density slightly deviated from the ground truth, our approach precisely predicts the future density, as illustrated in Figure~\ref{plot_eth2}.

\begin{table}[]
\caption{Results for the GroupNet+SSWNP model using various noise factor values during training. The noise factor ($\omega$) of 0.05 exhibits the best ADE/FDE values on the validation data.}
\label{tab:opt_noise}
\centering
\begin{tabular}{c|c|cccccccc}
\hline
\multirow{3}{*}{\begin{tabular}[c]{@{}c@{}}Model/ \\ Dataset\end{tabular}} &
  \multirow{3}{*}{\begin{tabular}[c]{@{}c@{}}Noise \\ Factor\end{tabular}} &
  \multicolumn{8}{c}{Validation Accuracy} \\ \cline{3-10} 
 &
   &
  \multicolumn{1}{c|}{ADE } &
  \multicolumn{1}{c}{FDE} \\ \cline{3-10} 
 &
   &
  
  \multicolumn{1}{c|}{4.0s} &
  4.0s \\ \hline
\multirow{4}{*}{\begin{tabular}[c]{@{}c@{}}GroupNet, \\ NBA\end{tabular}} &
  1 &
  \multicolumn{1}{c|}{0.908} &
  1.154 \\ \cline{2-10} 
 &
  0.1 &
  \multicolumn{1}{c|}{0.905} &

  1.131 \\ \cline{2-10} 
 &
  0.05 &
  \multicolumn{1}{c|}{\textbf{0.896}} &
  \textbf{1.130} \\ \cline{2-10} 
 &
  0 &
  \multicolumn{1}{c|}{1.13} &
  1.69 \\ \hline
\end{tabular}%
\end{table}

\begin{table*}[]
\caption{The results from experiments, which involved introducing noisy and clean environments in the trajectory sequence during testing, reveal that \textit{SSWNP} demonstrates resilience, whereas baseline models experience a significant performance decline. RD($\%$) represents the relative percent difference compared to the baseline.}
\label{tab:noise_env}
\centering
\begin{tabular}{c|c|c|cc|cc|c}
\hline
\multirow{2}{*}{Methods} &
  \multirow{2}{*}{Datasets} &
  \multirow{2}{*}{Environment} &
  \multicolumn{2}{c|}{Baseline} &
  \multicolumn{2}{c|}{Our} &
  \multirow{2}{*}{\begin{tabular}[c]{@{}c@{}}RD(\%)\\ (ADE/FDE)\end{tabular}} \\ \cline{4-7}
         &           &       & \multicolumn{1}{c|}{ADE}   & FDE   & \multicolumn{1}{c|}{ADE}   & FDE   &       \\ \hline
AutoBot  & TrajNet++ & Clean& \multicolumn{1}{c|}{0.128} &0.234& \multicolumn{1}{c|}{\textbf{0.091}} & \textbf{0.162} & \textbf{33.8/36.4} \\ \hline
AutoBot  & TrajNet++ & Noisy & \multicolumn{1}{c|}{0.301} & 0.469 & \multicolumn{1}{c|}{\textbf{0.134}} & \textbf{0.195} & \textbf{76.8/82.5} \\ \hline
GroupNet & NBA       & Clean & \multicolumn{1}{c|}{1.13} & 1.69 & \multicolumn{1}{c|}{\textbf{0.90}}  & \textbf{1.14}  & \textbf{22.7/38.9} \\ \hline
GroupNet & NBA       & Noisy & \multicolumn{1}{c|}{1.784} & 1.771 & \multicolumn{1}{c|}{\textbf{0.95}}  & \textbf{1.23}  & \textbf{61.0/36.1} \\ \hline
\end{tabular}%
\end{table*}

\subsection{Ablation Studies} \label{sec_abl}

\subsubsection{Significance of Spatial Consistency Module and Noise Prediction Module}
\label{ab_module}
We conducted experiments to validate the different modules of our approach. The results in Table~\ref{tab:self_task} show that including the proposed pretext tasks (Model B+SC+NP) improves trajectory prediction performance from 1.13/1.69 to 0.903/1.147 (ADE/FDE values), representing a relative percentage difference of 22.33/38.28\%. Furthermore, the (B+SC) model achieved a result of 1.018/1.362, which is lower than that of our (B+SC+NP) model.

\subsubsection{Choice of Noise Factor} 
\label{ab_noisefactor}
We conducted a study to evaluate the selection of the noise factor ($\omega$) for training our SSWNP. This noise factor is crucial as it regulates the impact of noise in repositioning spatial waypoints to generate diverse movement patterns in the data manifold. The results are outlined in Table~\ref{tab:opt_noise}; it is worth noting that the value of the noise factor may vary from dataset to dataset, and its determination could involve the use of cross-validation.

\subsubsection{Clean vs. Noisy Environment}
\label{ab_noisyenv}
We assessed the effectiveness of our approach in a noisy environment by introducing noise to both the baseline and our method during test time, then evaluating the predictions from each. The results are presented in Table~\ref{tab:noise_env}. On TrajNet++, the introduction of noise led to a deterioration in baseline method performance compared to our approach, with a reported difference (Relative percentage difference) of 76.8/82.5\% in ADE/FDE. Similarly, for the NBA dataset, we observed a difference of 61.0/36.1\% in the ADE/FDE values between our model and the baseline, indicating that our model performs significantly better in a noisy environment.

\section{Conclusion}
This work proposes a novel approach named SSWNP (Self-Supervised Waypoint Noise Prediction), consisting of spatial consistency and noise prediction modules. Our approach generates clean, noise-augmented views of historical trajectories observed over spatial waypoints. Subsequently, we enforce the trajectory prediction model to maintain spatial consistency between predictions derived from these two views. We also propose a novel pretext task focused on noise prediction within observed trajectories. This self-supervised task contributes to a deeper understanding of underlying representations in trajectory prediction, thereby enhancing the accuracy of future predictions. Experimental results show that incorporating SSWNP into the model learning process yields substantial performance improvements, even in noisy environments, when compared to baseline methods. This underscores the potential of our approach as a valuable complement to existing trajectory prediction techniques.

\bibliographystyle{IEEEtran}
\bibliography{ref}
\vfill

\end{document}